\documentclass[letterpaper, 10 pt, conference]{ieeeconf}  

\IEEEoverridecommandlockouts                              

\usepackage[noadjust]{cite}

\usepackage{booktabs}
\usepackage{float}
\usepackage{amsmath}
\usepackage{amsfonts}
\usepackage{graphicx}
\usepackage{subcaption}
\let\labelindent\relax
\usepackage{enumitem}
\usepackage[colorinlistoftodos,prependcaption,textsize=tiny]{todonotes}
\usepackage{gensymb}
\usepackage{hyperref}
\usepackage{lipsum}
\usepackage{balance}

\title{\LARGE \bf
Evaluating Robustness of Visual Representations for Object Assembly Task Requiring Spatio-Geometrical Reasoning
}

\author{Chahyon Ku$^{1}$, Carl Winge$^{1}$, Ryan Diaz$^{1}$, Wentao Yuan$^{2}$ and Karthik Desingh$^{1}$
\thanks{$^{1}$Minnesota Robotics Institute (MnRI) and Department of Computer Science and Engineering (CS\&E), University of Minnesota, Minneapolis, MN 55414 US.
{\tt\small(ku000045|winge134|diaz0329|kdesingh)@umn.edu}}%
\thanks{$^{2}$Paul. D Allen School of Computer Science and Engineering, University of Washington, Seattle, WA 98195 US. {\tt\small (wentaoy@cs.washington.edu)}}%
}

\let\oldtwocolumn\twocolumn
\renewcommand\twocolumn[1][]{%
    \oldtwocolumn[{#1}{
           \centering
           \includegraphics[width=0.85\textwidth]{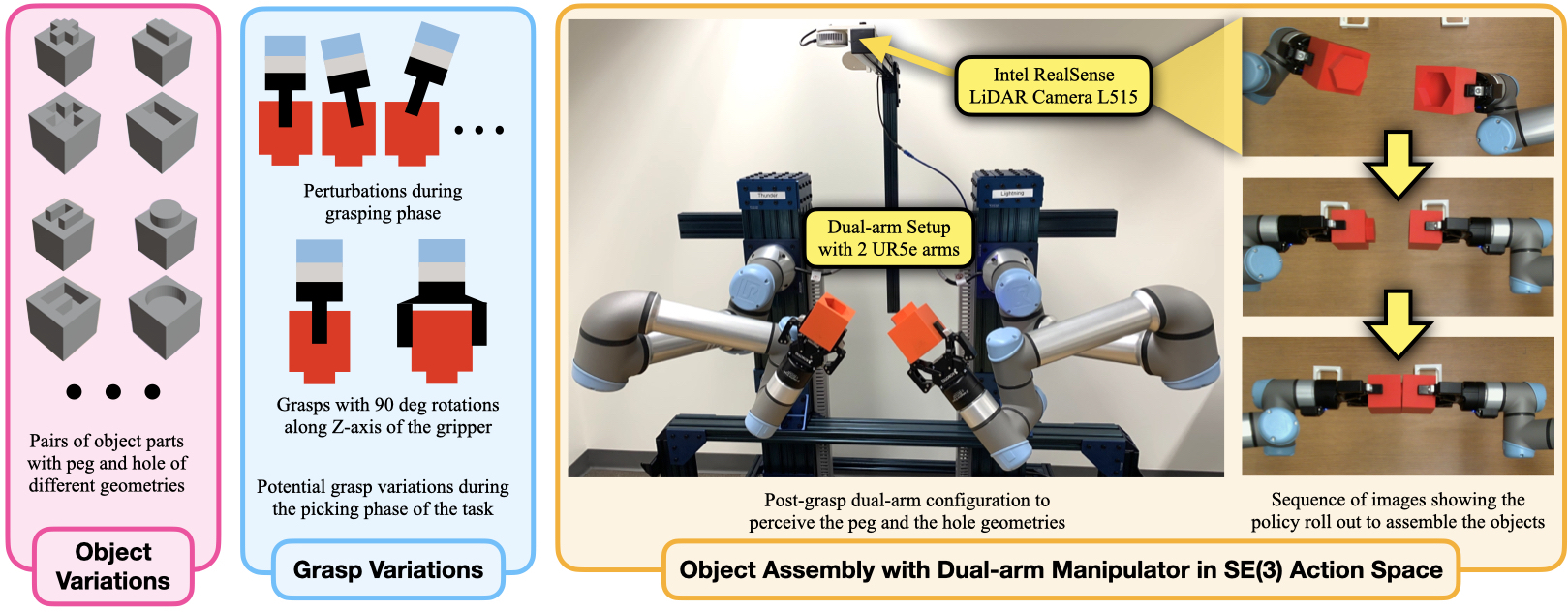}
           \captionof{figure}{\footnotesize{An overview of our benchmarking setup. Benchmarking robustness under object variations (left) and grasp variations (center) of visual policy learning methods on object assembly task with a dual-arm manipulator in SE(3) action space (right)} }
           \label{fig:teaser}
    }]
}

\usepackage{etoolbox}
\makeatletter
\patchcmd{\@makecaption}
  {\scshape}
  {}
  {}
  {}
\makeatother

\begin{document}
\maketitle
\begin{abstract}

This paper primarily focuses on evaluating and benchmarking the robustness of visual representations in the context of object assembly tasks. Specifically, it investigates the alignment and insertion of objects with geometrical extrusions and intrusions, commonly referred to as a peg-in-hole task. The accuracy required to detect and orient the peg and the hole geometry in SE(3) space for successful assembly poses significant challenges. Addressing this, we employ a general framework in visuomotor policy learning that utilizes visual pretraining models as vision encoders. Our study investigates the robustness of this framework when applied to a dual-arm manipulation setup, specifically to the grasp variations. Our quantitative analysis shows that existing pretrained models fail to capture the essential visual features necessary for this task. However, a visual encoder trained from scratch consistently outperforms the frozen pretrained models. Moreover, we discuss rotation representations and associated loss functions that substantially improve policy learning. We present a novel task scenario designed to evaluate the progress in visuomotor policy learning, with a specific focus on improving the robustness of intricate assembly tasks that require both geometrical and spatial reasoning. 
Videos, additional experiments, dataset, and code are available at \href{https://bit.ly/geometric-peg-in-hole}{https://bit.ly/geometric-peg-in-hole}.

\end{abstract}

\section{INTRODUCTION}
\label{sec:introduction}

Peg-in-hole assembly has been a longstanding and actively researched problem within the field of robotics as an essential component of industrial robotics that require reasoning over geometries, affordances, and contact models of objects. 
Prior works have predominantly focused on single-arm top-down peg-in-hole tasks, where the hole is attached to a fixture, while a robot arm picks and inserts the peg in top-down action space~\cite{park2017compliance, li2017human, azulay2022haptic, morgan2023towards,  song2016guidance, gao2021kpam, lu2022cfvs, van2018comparative}.
While these approaches effectively reduce the complexity of the task to focus on contact modelling and sub-millimeter-precision for industrial use cases, they make assumptions unfit for robot manipulation in more unstructured environments.
For robots to adequately manipulate household items with parts that fit together (e.g. containers and lids), we consider the following requirements to be essential.
To manipulate a diverse set of containers and lids the robot will encounter, the robot should be able to look at and reason about the intra-category variations of geometry (food containers vs. water bottles) as well as the objects' spatial (what is the relative transformation between parts?) and geometric (what parts fit together in what way?) relationship.
To operate in unstructured indoor environments, the robot may not assume objects are fixed to a flat surface but rather pick up both relevant parts and assemble them using two arms in full SE(3) action space.
The robot also may not have multiple external cameras giving a near-complete third-person-views of the task, the robot, and the objects.

Based on these requirements, we propose a novel geometric-peg-in-hole task, inspired by traditional peg-in-hole but adapted for object variations and unstructured environments, that evaluates the spatio-geometric reasoning capabilities of visuo-motor policies.
Our pegs and holes have geometric variations in the extrusions such as plus-sign, minus-sign, or pentagon, which require the robot to generalize over these intra-category variations without prior knowledge, as well as the different amounts of rotational variation required to put them together (see Fig. \ref{fig:object_dataset} with rotational symmetries).
The task starts out with the peg and hole in each gripper, but with randomized grasp noise to test the policy's ability to generalize over uncertainties during grasp estimation and execution (see Fig.~\ref{fig:teaser} center).
As both objects are held and controlled by the arms, the robot should not only reason about the current and goal SE(3) transformations between the objects, but also make use of all axes of SE(3) action space to insert the objects.
On top of the simulation environment created using PyBullet \cite{pybullet}, we also design a repeatable task setup in the real world, where the objects are always picked up and put down in the same position and orientation on a fixture, such that data collection and policy evaluation can be done without human intervention (see supplementary video and Fig.~\ref{fig:teaser} right).


Then, how can we train a robot to complete such a task?
End-to-end learning from demonstrations offers a flexible framework for training robust robots without the need to explicitly model object and task representations.
Taking wisdom from the recent success of pre-trained representations in natural language processing and computer vision, recent works such as R3M \cite{nair2022r3m} and MVP \cite{mvp} propose robotics-specific visual representations trained from large-scale datasets of ego-centric videos.
To assess the geometric reasoning capability of visual representations, we conduct a performance comparison. This comparison involves evaluating the performance of from-scratch trained baselines against various pre-trained representations in a image-to-action imitation learning setup.


In summary, our main contributions are:
\begin{itemize}[leftmargin=0.5cm]
\item A novel dual-arm geometric-peg-in-hole task designed to evaluate spatio-geometric reasoning capabilities of visual representations on 9 pairs of peg and hole objects with randomized grasp in rotation and translation of all axes.
\item A comprehensive analysis of eight visual representations, including two from-scratch encoders and six pre-trained models, is conducted to train policies within an imitation learning framework. This assessment quantitatively measures their success in handling increasingly challenging task (grasp) variations with two different imitation learning methods.
\item An evaluation of control representations (absolute vs. delta) and rotation representations (quaternion vs. Gram-Schmidt) in the context of our task learning.
\item A simulation environment, dataset, trained models, and evaluation code for others to benchmark and extend upon this task.

\end{itemize}

\section{RELATED WORKS}
\label{sec:related_work}

\subsection{Imitation Learning for Assembly Tasks}
End-to-end learning has demonstrated that robots can learn to control their actuators directly from raw sensor observations~\cite{deep_visuomotor, ghadirzadeh2017deep, duan2017one, huang2019neural}, avoiding the need for a brittle modular pipeline with explicit state estimation.
Traditionally, imitation learning works such as Form2Fit \cite{zakka2020form2fit} and Transporter \cite{zeng2020transporter} formulated assembly tasks as a supervised classification task of predicting SE(2) pick-and-place locations to train high-performing and data-efficient imitation learning policies.
These methods require a tabletop setup with a clear camera-view of the objects, stable insertion targets, and 2 dimensional action spaces.
More recently, Robomimic \cite{robomimic} proposed an imitation learning benchmark with SE(3) action space, and trained imitation learning policies for assembly-like tasks such as Tool Hang (insert hook onto base frame, hang tool on hook) and Square (place square nut on a rod).
While these tasks featured SE(3) action space and rotation of grasped objects, they focused on a single set of objects without geometric variation and on single-arm top-down insertion.
In this work, we propose a dual-arm SE(3) insertion task that evaluates the robot's ability to generalize over object variations and reason about SE(3) transformations between the parts.

\subsection{Pretraining for Robotics}
To avoid retraining the neural network for every task, more recent works have leveraged state-of-the-art visual pretrained models such as ResNet~\cite{resnet} pretrained on ImageNet~\cite{imagenet}, CLIP~\cite{clip}, MAE~\cite{mae}, and R3M~\cite{nair2022r3m} to obtain visual features to train a policy network. CLIP~\cite{clip} achieves state-of-the-art results in tasks like ImageNet classification by performing contrastive pretraining on a large-scale non-public image-caption dataset. R3M~\cite{nair2022r3m} utilizes time contrastive loss and video-language alignment to pretrain a ResNet~\cite{resnet} on egocentric videos sourced from the Ego4D dataset~\cite{grauman2022ego4d}. MAE~\cite{mae} achieves state-of-the-art transfer results in image classification by pretraining visual transformers~\cite{vit} on the ImageNet data. The pretraining process involves using masked input reconstruction, also known as masked autoencoding, as the supervision signal. MVP~\cite{mvp} utilizes masked autoencoding from MAE to pretrain vision transformers~\cite{vit} on egocentric videos sourced from a combination of datasets, such as Epic Kitchens~\cite{grauman2022ego4d}. In this work, we compare Non-pretrained ResNet-18 and ResNet-50 models with the above mentioned pretrained models except for MVP (does not provide ViT/B or ResNet-50 variants others commonly have).


\section{IMITATION LEARNING FRAMEWORK}
\label{sec:architecture}

\begin{figure*}[h]
     \centering
     \includegraphics[width=0.825\textwidth]{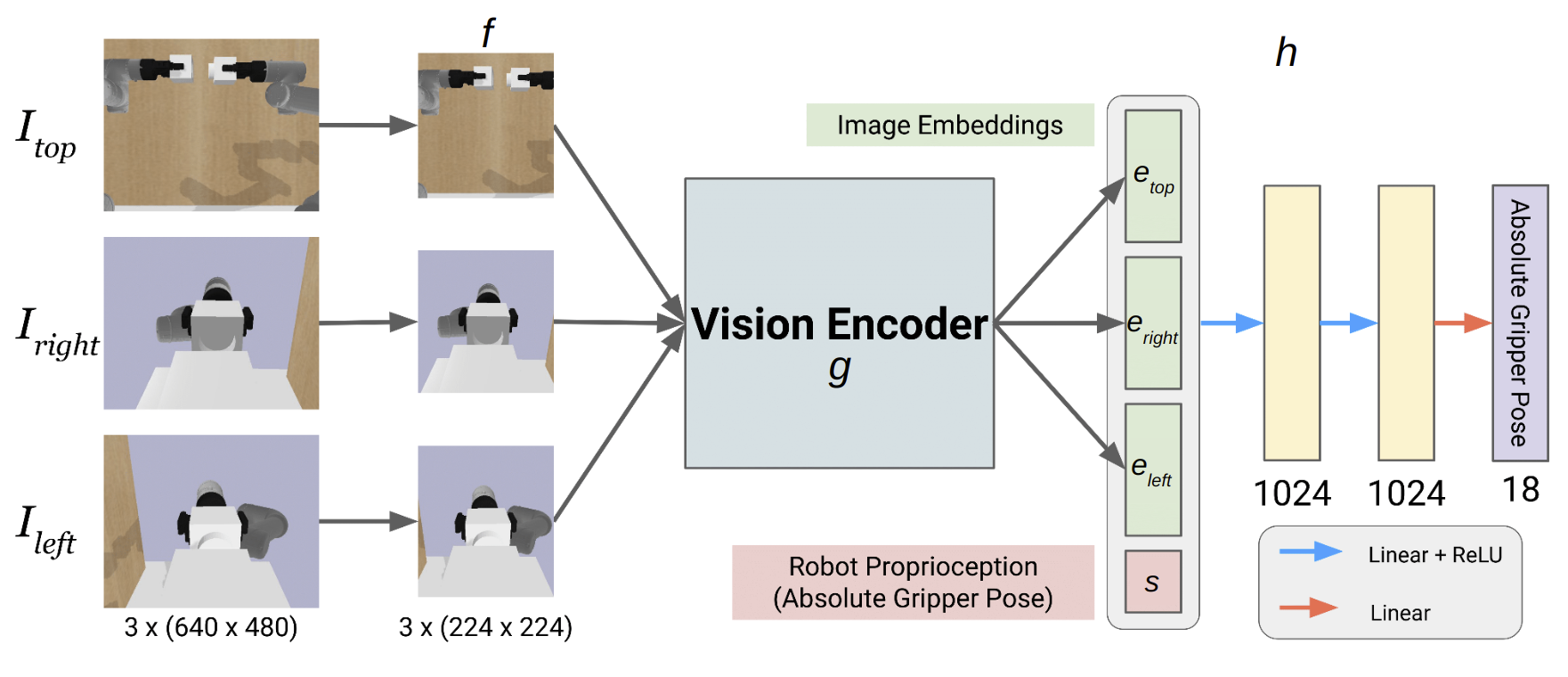}
     \caption{\footnotesize{Network architecture used in the imitation learning pipeline. The network receives 3 images as input, produces image embeddings for each image using a vision encoder of choice (pretrained or non-pretrained), concatenates them all with robot proprioception, and outputs the next action through an MLP head.}}
     \label{fig:architecture}
\end{figure*}
 
The goal of imitation learning is to train a policy $\pi: \mathcal{O} \rightarrow \mathcal{A}$ that maps all observations to an action that will progress the robot towards executing the task.
Our dataset $\mathcal{D} = \{(O_i, a_i)\}_{i=1,...N}$ consists of $N$ observation-action pairs. Each observation $O_i = ((I_{v})_{v \in V}, s)$ is a tuple of RGB images $I_{v}$ from view $v \in V$ and the current state of the robot $s \in \mathcal{S}$. Each action $a_i \in \mathcal{A}$ is the action the expert performs during demonstration. 

Our architecture shown in Fig.~\ref{fig:architecture} is inspired by imitation learning evaluation frameworks from Robomimic~\cite{robomimic}, R3M~\cite{nair2022r3m}, and MVP~\cite{mvp}. In our implementation, the policy $\pi_\theta$ consists of three parts: the image preprocessor $f$, the image encoder $g_\theta$, and the policy head $h_\phi$. The image preprocessor $f: \mathcal{I}^{640 \times 480} \rightarrow \mathcal{I}^{224 \times 224}$ crops and resizes the original RGB image to a consistant size for fair comparison of all vision encoder models (ViT-B/16 requires this size). The image encoder $g_{\theta}: \mathcal{I}^{224 \times 224} \rightarrow \mathbb{R}^D$ is a neural network that deterministically maps an RGB image $I_v$ to a $D$-dimensional image embedding $e_v$. The policy head $h_{\phi}: \mathbb{R}^D \times ... \times \mathbb{R}^D \times \mathcal{S} \rightarrow \mathcal{A}$ is a multi-layer perceptron on top of concatenated image embeddings $(e_v)_{v \in V}$ and robot state $s$ which produces the final output action $a$. In summary, $\pi_\theta(O) = h_{\phi}(g_{\theta}(f(I_{v_1})), ..., g_{\phi}(f(I_{v_{|V|}})), s) = \hat{a}$. We train either the parameters $\phi$ (frozen $g$) or both $\theta$ and $\phi$ (unfrozen $g$) by back-propagating the mean squared error loss $\mathcal{L} = MSE(a, \hat{a})$. 

We choose $\mathcal{S} \subseteq SE(3) \times SE(3)$ and $\mathcal{A} \subseteq SE(3) \times SE(3)$ to be absolute gripper poses of both arms. For each arm, there are 3 values representing xyz position and 6 values representing the first two columns of the rotation matrix \cite{continuity_rotation}, so we represent both as a 18 dimensional vector. The reason for this choice is explained in Sec. \ref{sec:experiments}.



\section{TASK SETUP}
\label{sec:task_setup}
\subsection{Objects}
\begin{figure}[h!]
\centering
\includegraphics[width=0.75\columnwidth]{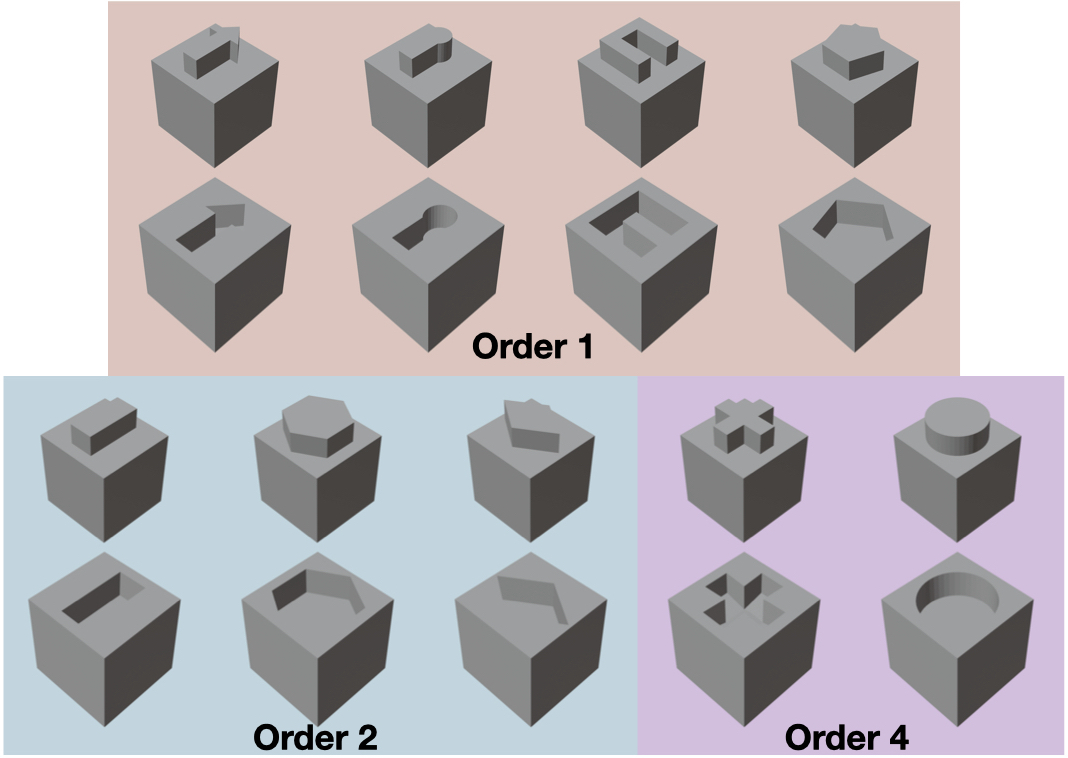}
\caption{\footnotesize{9 peg and hole pairs with various shapes, such as \textit{arrow}, \textit{key}, \textit{U}, \textit{pentagon}, \textit{minus-sign}, \textit{hexagon}, \textit{diamond}, \textit{plus-sign}, and \textit{circle}, grouped into categories with rotational symmetries of orders 1, 2, and 4.}}
\label{fig:object_dataset}
\vspace*{-5mm}
\end{figure}

We generate 3D models of nine peg and hole object pairs using Blender~\cite{blender} (Fig.~\ref{fig:object_dataset}).
Each object pair has one object (the ``peg'') with an extrusion of a specific geometrical shape, with the other object (the ``hole'') having an equivalently-shaped hole. The objects are categorized based on their rotational symmetries (i.e. of the orders 1, 2, and 4). When rotating the object by \{$0\degree, 90\degree, 180\degree, 270\degree$\}, order 1 objects have 4 unique orientations (see Fig. \ref{fig:object_dataset}, top row), order 2 objects have 2 unique orientations, and order 4 objects have 1 unique orientation. 
The objects used in the real-world environment are 3D printed directly from the generated 3D models.

\subsection{Simulation Setup}
\begin{figure*}[h!]
\centering
\includegraphics[width=0.9\textwidth]{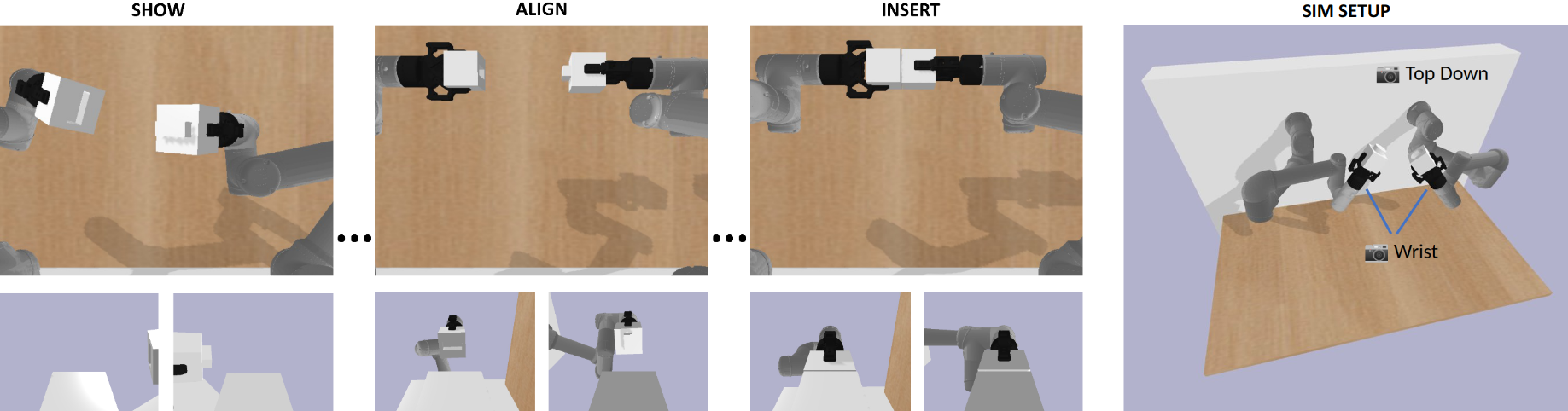}
\caption{\footnotesize{Task execution sequence in the simulation setup with top-view and left and right wrist-views.}}
\label{fig:task_exec_sequence_sim}
\end{figure*}

The PyBullet \cite{pybullet} simulation environment (see Fig.~\ref{fig:task_exec_sequence_sim}) is used for data collection and imitation learning evaluation.
It consists of the calibrated clone of our real world setup including two 6-DoF robot arms (Universal Robots UR5e), with two parallel jaw grippers (2F-85 Robotiq grippers) and one top-down view RGB camera (Intel RealSense LiDAR Camera L515 in the real-world).
Additionally, two wrist RGB cameras are added in the simulation setup to experiment if multiple views help with the policy learning.

\begin{figure}[h!]
\centering
\includegraphics[width=1.0\columnwidth]{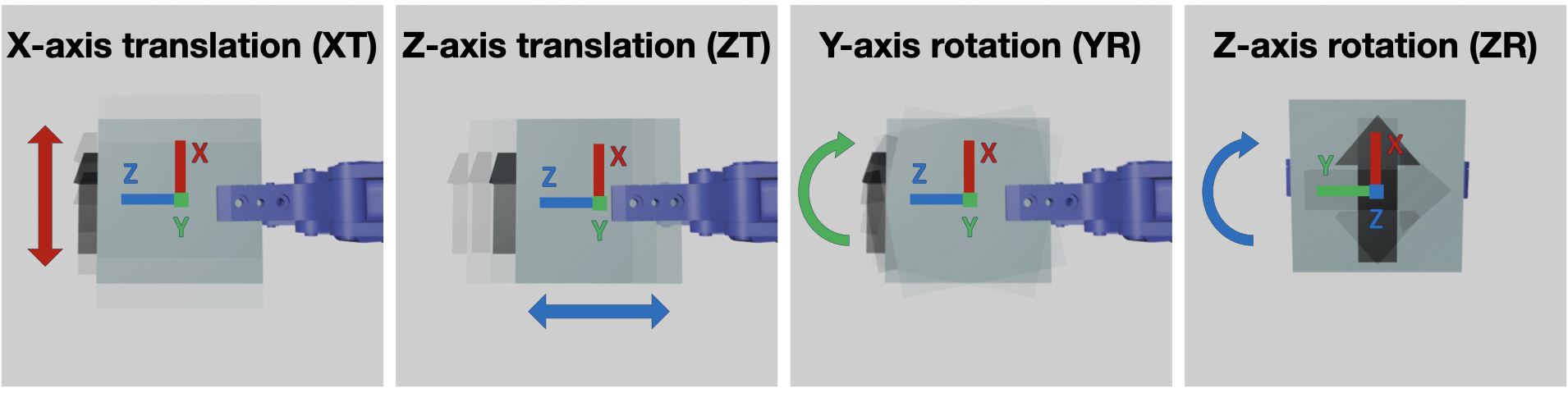}
\caption{\footnotesize{Possible variations in grasping the objects.}}
\label{fig:grasp_variations}
\vspace*{-5mm}
\end{figure}



\textbf{Task Initialization:}
In the beginning of the episode, peg and hole are randomly held in the gripper based on the specific task variation.
As shown in Fig.~\ref{fig:grasp_variations}, the in-hand pose of the cap and bottle are randomized from the base pose in 4 ways:
XT is the noise in the X-axis Translation of the gripper, uniformly sampled from [-0.01m, 0.01m].
ZT is the noise in the Z-axis Translation of the gripper, uniformly sampled from [-0.01m, 0.01m].
YR is the noise in the Y-axis Rotation of the gripper, uniformly sampled from [-11.25$^\circ$, 11.25$^\circ$].
ZR is the random Z-axis Rotation of the gripper, uniformly sampled from \{0$^\circ$, 90$^\circ$, 180$^\circ$, 270$^\circ$\}.
Note that ZR is not noise but rather a discrete set of rotations that requires the model to conduct geometric reasoning, whereas XT, ZT, and YR are noisy perturbations.

\textbf{Scripted Expert Demonstrations:} 
The task starts in the ``show'' state (see Fig.~\ref{fig:task_exec_sequence_sim}): a fixed joint configuration that points both grippers towards the camera.
With the objects in hand, both arms simultaneously move towards the ``align'' pose (see Fig.~\ref{fig:task_exec_sequence_sim}), where the objects are perfectly aligned with each other.
Afterwards, the arms make a simple linear cartesian move towards the ``insert'' pose (see Fig.~\ref{fig:task_exec_sequence_sim}), where the objects are put together.

\textbf{Demonstration Statistics:} Roughly 44\% of the demonstrations have order 1 objects, 33\% have order 2 objects, and 22\% have order 4 objects due to the distribution of possible shapes in each category (see Fig. \ref{fig:object_dataset}).
Each demonstration contains a minimum of 10 frames and a maximum of 40 frames.
The task is considered successful if the relative transform of the peg and the hole are apart within a margin of 1cm in translation and $5^\circ$ in orientation from the expected ground-truth relative transform of the peg and the hole.
While this is a relaxation of the precision required for industrial peg-in-hole tasks, it still poses a significant challenges to vision-based systems, as SOTA algorithms such as GDR-Net \cite{Wang_2021_GDRN} are evaluated on 2cm accuracy for tabletop object pose estimation.



\subsection{Real World Setup}
The real environment (as seen in Fig.~\ref{fig:teaser} right) is identical to the simulation environment with a few additions to allow for automatic data collection and evaluation.
Since objects cannot simply be spawned into the real-life grippers like in simulation, we create a picking phase before the episode and a placing phase after the episode.
After the blocks are picked, the robot moves to a predefined joint state to show the blocks to the camera.
From there, the robot runs either the scripted demonstration or rollout of the trained policy as in the simulation.
After the execution, the system always places the peg and hole in a fixed pose such that the extrusion and intrusion are face-down from the top-view inside 2 3D-printed fixtures to ensure precise placement.
Thus, the system can always assume that the peg and hole are picked up from the same pose and can vary the gripper-object transformation as needed for each task variation.
For more details on the real world setup, refer to the supplementary video.


\section{EXPERIMENTS}
\label{sec:experiments}
\subsection{Experimental Setup}\label{sec:exp_setup}

\textbf{Task Variations:}
We compare performance on the four grasp variations (XT, ZT, YR, ZR) shown in Fig.~\ref{fig:grasp_variations} and their combinations XTZR, ZTZR, YRZR, and XZTYZR (denoting all variations). The ZR variation requires the model to perform geometric-spatial reasoning to succeed, whereas the other variations require the model to be robust to perturbations in the grasping. Hence, complex variation compositions involving ZR requires the model to do geometric-spatial reasoning to perform the assembly task successfully.

\textbf{Model Variations:}
We compare the performance of 2 randomly initialized image encoders and 6 pretrained frozen image encoders (3 ResNet~\cite{resnet} models and 3 vision transformer~\cite{vit} models).
Non-pretrained ResNet-18 and ResNet-50 are randomly initialized and jointly trained with the policy head.
ImageNet ResNet-50, R3M ResNet-50, and CLIP ResNet-50 are frozen ResNet-50s initialized with ImageNet~\cite{imagenet}, R3M \cite{nair2022r3m}, and CLIP~\cite{clip} weights.
ImageNet ViT-B/16, CLIP ViT-B/16, and MAE ViT-B/16 are frozen ViT-B/16s initialized with ImageNet, CLIP, and MAE~\cite{mae} weights.

We use ResNet-50 and ViT-B/16 variants of the pretrained models, because they are the only versions available for all above-mentioned pretraining schemes and have been used for imitation learning in previous works~\cite{nair2022r3m, mvp}.

\textbf{Object Set Variations:}
As shown in Fig. \ref{fig:object_dataset}, we have objects with rotational symmetries of Order-1, Order-2, and Order-4.
Order-all denotes the union of Order-1, Order-2, and Order-4, which is used for training all models.


\subsection{Simulation Results}
In this section, we will compare the performance of models trained and tested with simulation.
Unless otherwise indicated, we use Non-pretrained ResNet-18.
\textbf{Note:} All models in the simulation experiments are trained with datasets containing all 9 objects from Order-1, Order-2, and Order-4 to test the models' capability to generalize over geometric variations.

\textbf{Comparison of Visual Representations:} All visual representation, pretrained or non-pretrained, are mostly successful in task variants with just grasp noise perturbations, with success rates of 0.900 or above.
However, we observe that the non-pretrained models perform best on ZR that requires geometric-spatial reasoning, with average 0.800 success rates, compared to the best performing pretrained models (CLIP) with average 0.600 success rate (TABLE \ref{fig:all_models_1_variation}).

\begin{table}[h!]
\centering
    \begin{tabular}{lllll}
    \toprule
   & \multicolumn{1}{c}{\textbf{XT}} & \multicolumn{1}{c}{\textbf{ZT}} & \multicolumn{1}{c}{\textbf{YR}} & \multicolumn{1}{c}{\textbf{ZR}} \\
   \midrule
    Non-pretrained ResNet-18 & \textbf{1.000} & \textbf{1.000} & \textbf{1.000} & \textbf{0.775}\\
    Non-pretrained ResNet-50 & \textbf{1.000} & \textbf{1.000} & \textbf{1.000} & \textbf{0.825}                     \\
    ImageNet ResNet-50       & 1.000 & 1.000 & 0.925                     & 0.425                     \\
    R3M ResNet-50            & 0.950                        & 1.000 & 1.000 & 0.275                     \\
    CLIP ResNet-50           & 1.000 & 1.000 & 0.975                     & 0.625                     \\
    ImageNet ViT-base        & 0.950                        & 1.000 & 0.975                     & 0.450                      \\
    CLIP ViT-base            & 1.000 & 1.000 & 0.900                       & 0.575                     \\
    MAE ViT-base             & 1.000 & 1.000 & 0.925                     & 0.350                \\ 
    \bottomrule
    \end{tabular}
     \caption{\footnotesize{Success rates of all visual representations trained with 100 demonstrations of indicated task variation. Non-pretrained ResNets clearly outperform pretrained models on ZR.}}
     \label{fig:all_models_1_variation}
\end{table}

The performance on combinations of grasp variations (XTZR, ZTZR, and YRZR) is much worse when trained using the same amount of data (100 episodes). Success rates of the models range from \textbf{0.400} to \textbf{0.600} for Non-pretrained ResNet-18, the best performing model.
We hypothesize that the amount of data (100 episodes) is not sufficient enough to capture the features to successfully complete the trajectory, and confirm that increasing the amount of demonstrations drastically improves performance from 0.38 average success rate to 0.65 and 0.70 average success rate (TABLE~\ref{fig:all_models_all_variation}).

\begin{table}[h!]
\centering
\begin{tabular}{lllll}
\toprule
 & \multicolumn{1}{c}{\textbf{XTZR}} & \multicolumn{1}{c}{\textbf{ZTZR}} & \multicolumn{1}{c}{\textbf{YRZR}} & \multicolumn{1}{c}{\textbf{XZTYZR}} \\ \midrule
Non-pretrained ResNet-18 & \textbf{0.825} & \textbf{0.825} & \textbf{0.675} & \textbf{0.275} \\
Non-pretrained ResNet-50 & 0.425 & 0.775 & 0.300 & 0.075 \\
ImageNet ResNet-50 & 0.225 & 0.225 & 0.175 & 0.050 \\
R3M ResNet-50 & 0.150 & 0.275 & 0.05 & 0.050 \\
CLIP ResNet-50 & 0.500 & 0.575 & 0.250 & 0.150 \\
ImageNet ViT-base & 0.150 & 0.300 & 0.225 & 0.025 \\
CLIP ViT-base & 0.300 & 0.250 & 0.200 & 0.050 \\
MAE ViT-base & 0.375 & 0.25 & 0.175 & 0.050 \\ \bottomrule
\end{tabular}
 \caption{\footnotesize{Success rates of all visual representations trained with 1000 demonstrations of indicated task variation using all objects.}}
 \label{fig:all_models_all_variation}
\end{table}


\textbf{Comparison of Task and Object Variations:} We provide a more fine-grained evaluation of our best model (Non-pretrained ResNet-18) by running 3 different evaluation runs of 40 randomized episodes and reporting the mean and standard deviation over the 3 runs (TABLE~\ref{fig:task_object_variation}). ``All'' denotes the average of all evaluation runs above (total 360 episodes). These models are trained on 1000 demonstrations, which includes all shapes, with proprioception and 3 views (top + 2 wrist cameras).

\begin{table}[h!]
\centering
\begin{tabular}{lllll}
\toprule
\multicolumn{1}{c}{\textbf{Objects}} & \multicolumn{1}{c}{\textbf{XTZR}} & \multicolumn{1}{c}{\textbf{ZTZR}} & \multicolumn{1}{c}{\textbf{YZR}} & \multicolumn{1}{c}{\textbf{XZTYZR}} \\ \midrule
circle & 0.85$\pm$0.07 & 1.00$\pm$0.00 & 0.83$\pm$0.05 & 0.43$\pm$0.07 \\
plus & 0.93$\pm$0.04 & 1.00$\pm$0.00 & 0.77$\pm$0.01 & 0.38$\pm$0.04 \\
minus & 0.80$\pm$0.03 & 0.98$\pm$0.00 & 0.44$\pm$0.10 & 0.33$\pm$0.10 \\
diamond & 0.77$\pm$0.06 & 1.00$\pm$0.00 & 0.33$\pm$0.08 & 0.34$\pm$0.08 \\
hexagon & 0.71$\pm$0.08 & 1.00$\pm$0.00 & 0.38$\pm$0.08 & 0.30$\pm$0.08 \\
u & 0.37$\pm$0.08 & 0.54$\pm$0.06 & 0.12$\pm$0.06 & 0.17$\pm$0.03 \\
pentagon & 0.34$\pm$0.10 & 0.56$\pm$0.04 & 0.10$\pm$0.07 & 0.18$\pm$0.07 \\
arrow & 0.38$\pm$0.07 & 0.66$\pm$0.08 & 0.18$\pm$0.03 & 0.17$\pm$0.05 \\
key & 0.38$\pm$0.07 & 0.66$\pm$0.08 & 0.17$\pm$0.04 & 0.19$\pm$0.05 \\
all & 0.61$\pm$0.02 & 0.82$\pm$0.03 & 0.37$\pm$0.05 & 0.28$\pm$0.03 \\ \bottomrule
\end{tabular}
\caption{\footnotesize{Success rates of Non-pretrained ResNet-18 trained on 1000 demonstrations including all objects. Mean and standard deviations over 3 different evaluations of 40 randomized rollouts.}}
\label{fig:task_object_variation}
\vspace*{-5mm}
\end{table}

We observe that the success rates are mostly consistent when evaluated on a completely new set of 40 randomized episodes, with a standard deviation falling around 0.05. We observe that while the success rate of rotation groups (Order-4: circle and plus, Order-2: minus, diamond, and hexagon, Order-1: u, pentagon, and arrow) decline consistently with increased degrees of symmetry. However, there is no clear trend in the order of performance for objects inside each group.

\textbf{Comparison of Imitation Learning Methods:} We compare our original setup, BC-MLP with various image encoders, against BC-RNN [1], a popular SOTA BC baseline that is shown to improve performance over BC-MLP. Instead of a MLP of layer sizes [1024, 1024, 18] on top of the image encoder, BC-RNN has two LSTM layers of hidden sizes [1024, 1024] and output size 18 that fuses image embeddings and proprioception from 10 most recent time frames as implemented in the original paper [1]. The tabulated results show success rates evaluated over 40 rollouts. These models are all trained on 1000 demonstrations with proprioception and 3 views (top + 2 wrist cameras).

\begin{table}[h!]
\centering
\begin{tabular}{lllll}
\toprule
\multicolumn{1}{c}{\textbf{Models}}& \multicolumn{1}{c}{\textbf{XTZR}} & \multicolumn{1}{c}{\textbf{ZTZR}} & \multicolumn{1}{c}{\textbf{YRZR}} & \multicolumn{1}{c}{\textbf{XZTYZR}} \\
\midrule
ResNet-18 MLP & \textbf{0.825} & \textbf{0.825} & \textbf{0.675} & \textbf{0.275} \\
ResNet-18 RNN & 0.525 & 0.300 & 0.350 & 0.075 \\
CLIP ResNet-50 MLP & 0.500 & 0.575 & 0.250 & 0.150 \\
CLIP ResNet-50 RNN & 0.300 & 0.325 & 0.150 & 0.200 \\
CLIP ViT-base MLP & 0.300 & 0.250 & 0.200 & 0.050 \\
CLIP ViT-base RNN & 0.150 & 0.350 & 0.025 & 0.025 \\
\bottomrule
\end{tabular}
\caption{\footnotesize{Success rates of Non-pretrained ResNet-18, CLIP ResNet-50, and CLIP ViT-Base with MLP vs LSTM action decoders, trained on 1000 demonstrations including all objects.}}%
\end{table}

\begin{table*}[t!]
\centering
\begin{tabular}{llllllllll}
\toprule
& \multicolumn{1}{c}{\textbf{Shape}} & \multicolumn{1}{c}{\textbf{XT}} & \multicolumn{1}{c}{\textbf{ZT}} & \multicolumn{1}{c}{\textbf{YR}} & \multicolumn{1}{c}{\textbf{ZR}} & \multicolumn{1}{c}{\textbf{XTZR}} & \multicolumn{1}{c}{\textbf{ZTZR}} & \multicolumn{1}{c}{\textbf{YRZR}} & \multicolumn{1}{c}{\textbf{XZTYZR}} \\
\midrule
 & plus & 0.9 & 1.0 & 0.7 & 1.0 & 0.6 & 0.7 & 0.2 & 0.3 \\
\textbf{One model per shape} & minus & 1.0 & 1.0 & 1.0 & 0.7 & 0.4 & 0.5 & 0.3 & 0.0 \\
 & key & 1.0 & 1.0 & 0.5 & 0.0 & 0.0 & 0.3 & 0.0 & 0.1 \\
 \midrule
 & plus & 0.3 & 0.2 & 0.2 & 0.9 & 0.2 & 0.5 & 0.1 & 0.0 \\
\textbf{One model for all shapes} & minus & 0.5 & 0.2 & 0.1 & 0.5 & 0.2 & 0.6 & 0.0 & 0.0 \\
 & key & 0.3 & 0.0 & 0.2 & 0.3 & 0.1 & 0.3 & 0.1 & 0.0 \\
 \bottomrule
\end{tabular}
\caption{\footnotesize{Real world results for one model per shape vs. one model for all shapes, trained from 10 demonstrations for each object}}
\label{fig:real_world}
\end{table*}

\begin{table}[t!]
\centering
\begin{tabular}{lllll}
\toprule
\textbf{Rotation-Loss} & \multicolumn{1}{l}{\textbf{XTZR}} & \multicolumn{1}{l}{\textbf{ZTZR}} & \multicolumn{1}{l}{\textbf{YRZR}} & \multicolumn{1}{l}{\textbf{XZTYZR}} \\ \midrule
Quaternion - MSE & 0.275 & 0.250 & 0.000 & 0.000 \\
Quaternion - Frobenius & 0.225 & 0.375 & 0.100 & 0.000 \\
6D - Frobenius & 0.575 & 0.625 & 0.200 & 0.125 \\
6D - MSE & 0.825 & 0.825 & 0.675 & 0.275 \\ \bottomrule
\end{tabular}
\caption{\footnotesize{Success rates of the Non-pretrained ResNet-18 using a combination of a certain rotation representation (either as a quaternion or as a 6d rotation matrix completed by Gram-Schmidt orthonormalization).}}
\label{fig:rotations}
\vspace*{-5mm}
\end{table}

Contrary to our expectations, we observe that the performance is worse for BC-RNN on all tasks and models we ran these experiments on: Non-pretrained ResNet-18, CLIP ResNet-50, and CLIP ViT-base. We hypothesize that BC-RNN cannot use the history as well as it performed in Robomimic \cite{robomimic}, because BC-RNN performs better on longer tasks while our tasks are relatively short. To be more specific, the 5 tasks in the RoboMimic have average episode lengths of 48 (Lift), 116 (Can), 151 (Square), 469 (Transport), and 480 (Tool Hang), where the performance of BC-RNN improved for only the longer three tasks. In comparison to their longer tasks, our tasks average around 10 to 40 frames. While ALOHA \cite{aloha}, another recent work also proposes improvements in temporal smoothing and consistency through use of transformers, we do not conduct experiment with their setup assuming insignificant improvements from the same reasons.
\\

\textbf{Comparison of Action Representation and Rotation Regression Targets:}
The choice of action and rotation representations was crucial for learning the policy as described in TABLE~\ref{fig:rotations}.
First, we observed that position control of the end-effector poses was the best action representation for the task, as velocity control was too sensitive to the step size of the action and was completely unable to correctly model trajectories that change sharply in the velocity space.
Second, we observed that using 6D representations for 3D rotations proposed in \cite{continuity_rotation}, which regresses the first 2 columns of the rotation matrix and performs Gram-Schmidt orthonormalization to produce the full rotation matrix, greatly outperformed the standard quaternion representation.


\subsection{Real World Results}


We conduct real world experiments to verify if (1) our observations from the simulation experiments are consistent with the real world (2) our setup can generalize with fewer demonstrations on the real robot. We experiment with a setup that is more similar to our simulation setup by training a single model to generalize over all the shapes. We calculate the success rates by counting ``smooth insertions," which does not include forcing the objects together such that objects slip into each other even when they are not aligned. In other words, we do not count unintended insertions that did not properly align the objects towards success. Additionally, we also experiment with a setup that has one model per shape. 

We observe that one model for all shapes (TABLE~\ref{fig:real_world} bottom) perform worse than models trained and evaluated in simulation, because of fewer number of demonstrations in the real world. In this setup, we were not able to draw conclusive observations on the exact trend of performance, due to high variance in success rates likely induced from limited number of train demonstrations (30 vs. 100/1000/10000) and evaluation runs (10 vs. 40). It is also worth noting that the real world setup does not have wrist cameras, adding to the performance drop compared to simulated experiments. In our additional experiments where we train one model per shape (TABLE~\ref{fig:real_world} top), we notice considerable improvements in success rates, suggesting that more demonstrations are needed to generalize over multiple shapes. However, the general trend of performance is consistent with our simulation setup: (a) objects with more degrees of symmetry perform better (b) tasks with simpler variations perform better than combinations (e.g. ZR performs better than XTZR, ZTZR, YRZR, and XZTYZR for all shapes except the key).

\subsection{Additional Details \& Analysis}

Apart from the core experiments mentioned above, we also provide additional details (Object Details, Training Hyperparameters) and experiments on the website\footnote{Project webpage: \href{https://bit.ly/geometric-peg-in-hole}{https://bit.ly/geometric-peg-in-hole}}:

\noindent \textbf{Frozen vs. Finetuning:} We conduct additional experiments by fine-tuning the pretrained image encoders with the action decoder to conclude that (a) Pretrained ResNet-50 models perform better when trained unfrozen, but still does not outperform from-scratch models and (b) Pretrained ViT-B/16 models are more unstable when trained unfrozen and often performs worse than the frozen counterpart.

\noindent \textbf{Data Efficiency of Models:} We compare the performane of all visual representations trained with 100, 1000, and 10000 training episodes to observe that the non-pretrained models consistently improves with more data while frozen models saturate in performance at 1000 demos due to the limited capacity of MLP action decoders.

\noindent \textbf{With vs. Without Proprioception:} We provide analysis of models trained without proprioception to isolate the effect of visual representations and observe that proprioception is crucial for task variations with Z rotation because the policy network benefits from explicityly knowing the robot's current state to accurately step toward the task goal.

\noindent \textbf{Colored Extrusions for Ease of Recognition}: We compare the performance of models trained on colored objects with more contrast between the base and the extrusion.

\section{CONCLUSION}
\label{sec:limitations}
In this paper, we propose a novel dual-arm geometric-peg-in-hole task and evaluate the spatio-gemoetric reasoning capabilities of various visual representations through an imitation learning framework.
We observe that a non-pretrained visual encoder with MLP policy decoder were most effective in completing the task across all variations.
For future work, we plan to explore data-driven contact modelling by incorporating force feedback while also extending the geometric variation of objects to more realistic objects.
Furthermore, we hypothesize that advanced imitation learning methods such as DiffusionPolicy \cite{diffusion_policy}, which is shown to perform well in multimodal learning tasks, may significantly improve performance on our task and leave it for future work.

\section*{ACKNOWLEDGMENT}

We thank Bahaa Aldeeb and Alireza Rezazadeh for providing helpful feedback on our initial draft and all other members of the Robotics: Perception and Manipulation (RPM) Lab for their insightful discussions. This project is partially funded by the URS and UROP Program at the University of Minnesota and the MnRI Seed Grant from the Minnesota Robotics Institute.





\end{document}